\documentclass[10pt,twocolumn,letterpaper]{article}

\usepackage{cvpr}

\usepackage[utf8]{inputenc} 
\usepackage[T1]{fontenc}    
\usepackage{url}            
\usepackage{booktabs}       
\usepackage{amsfonts}       
\usepackage{nicefrac}       
\usepackage{microtype}      
\usepackage{xcolor}         
\usepackage{graphicx}
\usepackage{amsmath}
\usepackage{soul}
\definecolor{highlightcolor}{rgb}{0.9, 0.9, 0.9}
\sethlcolor{highlightcolor}

\definecolor{cvprblue}{rgb}{0.21,0.49,0.74}
\usepackage[pagebackref,breaklinks,colorlinks,allcolors=cvprblue]{hyperref}

\title{Selfish Evolution: Making Discoveries in Extreme Label Noise with the Help of Overfitting Dynamics}

\author{
  Nima Sedaghat\textsuperscript{1,2}\thanks{\texttt{nimaseda@uw.edu}} \quad
  Tanawan Chatchadanoraset\textsuperscript{1,2} \quad
  Colin Orion Chandler\textsuperscript{1,2} \\
  Ashish Mahabal\textsuperscript{3} \quad
  Maryam Eslami\textsuperscript{4} \\\\[1ex]
  \textsuperscript{1}Department of Astronomy, University of Washington, Seattle, WA\\
  \textsuperscript{2}Raw Data Speaks, Seattle, WA\\
  \textsuperscript{3}California Institute of Technology (Caltech), Pasadena, CA\\
  \textsuperscript{4}Azad University, Tehran, Iran
}

\begin{document}

\maketitle

\begin{abstract}
Motivated by the scarcity of proper labels in an astrophysical application, we have developed a novel technique, called Selfish Evolution, which allows for the detection and correction of corrupted labels in a weakly supervised fashion.
Unlike methods based on early stopping, we let the model train on the noisy dataset. Only then do we intervene and allow the model to overfit to individual samples. The ``evolution'' of the model during this process reveals patterns with enough information about the noisiness of the label, as well as its correct version. We train a secondary network on these spatiotemporal ``evolution cubes'' to correct potentially corrupted labels. 
We incorporate the technique in a closed-loop fashion, allowing for automatic convergence towards a mostly clean dataset, without presumptions about the state of the network in which we intervene.
We evaluate on the main task of the Supernova-hunting dataset but also demonstrate efficiency on the more standard MNIST dataset.
\end{abstract}

\begin{figure}[h]
  \centering
  \scalebox{-1}[1]{\includegraphics[width=0.5\textwidth]{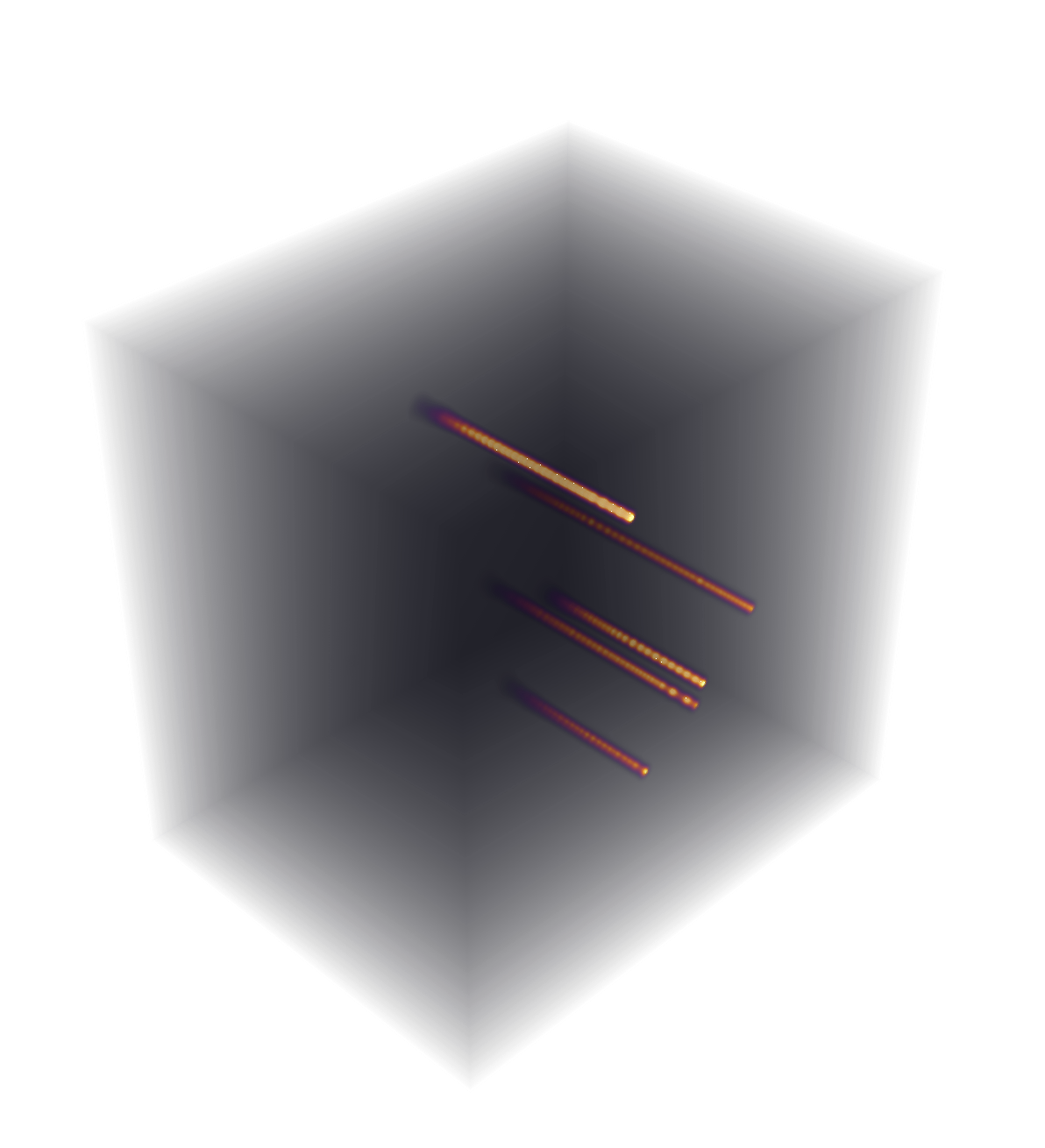}}
  \caption{Capturing the evolution of the model output during individual overfitting processes results in data volumes encapsulating a good amount of information about the presence of label noise, and potentially the noise-free label. Here we depict an exemplar ``evolution cube'' for a special approach to the task of supernova detection in which labels are 2D images. The third axis, representing the evolution steps, is aligned with the (dis)appearing objects in the above illustration.}
  \label{fig:teaser}
\end{figure}

\section{Introduction}

Deep learning and computer vision techniques have made significant inroads in various domains of science, including astronomy, where they are used to enhance data analysis and discovery processes \cite{chandler2024active, chandler2024ai, sedaghat20242016}. In the field of astronomy, one prominent application is the detection of celestial phenomena such as supernovae. By leveraging deep learning models, astronomers can analyze vast amounts of astronomical data efficiently and accurately, facilitating the identification of these explosive events. However, coming up with high-quality ground truth in the target, real domain, is extremely more difficult than typical earthly vision applications.

Tackling this issue is of high importance, as the scarcity of good labels not only impedes the model's ability to learn and generalize well but also the missed samples in the training set are potentially interesting objects. Specifically, a missed supernova in the training set, apart from contaminating the training process, may mean a missed, important discovery.

The state-of-the-art in the task of supernova detection is an image-generating approach called TransiNet---\cite{sedaghat2018effective}. The method generates images in which it tries to ``paint'' the detections on a blank canvas. 
In practical scenarios, there exists a high number of undetected true objects in the ground truth data, which substantially hinders the training process---\cite{mahabal2019machine,sedaghat2023deep,sedaghat2023report, sedaghat2024realbogus}. Due to the pixel-wise nature of the method, each missed object is virtually more than a single missed object: each pixel belonging to the missed object contributes to training the network on wrong labels.

In this work, we propose a method for detecting \textit{and} recovering these missed discoveries. We cast the problem as one of label noise, where the noise presents itself as a false negative: a supernova that has occurred in the past but has not been discovered yet. We extract subtle information out of the model dynamics while it is overfitting to each sample to get hints about the noisiness of the samples.

Label noise has a wide and well-studied body of literature---\cite{Wei2021LearningWN,song2022learning}. This aspect of machine learning research emphasizes the impact of incorrect labels on model performance, highlighting the need for robust techniques to mitigate its effects. 
Most existing studies often focus on developing algorithms that can withstand noisy labels, whether it is by dropping bad labels or weighting good labels---\cite{Frenay2014} and \cite{Algan2020}. As a result, a new sub-field under the name  Learning with Noisy Labels, LNL, has emerged. 
The field focuses on the development of models capable of effectively learning from datasets contaminated with label noise. 

Research in the area of LNL can be broadly categorized into two main approaches: robust algorithms and noise detection strategies. Robust algorithms are designed to enhance the resilience of the learning process without directly addressing the noise in individual data instances. These methods incorporate specific mechanisms to ensure that neural networks can be trained effectively despite the presence of label noise \cite{frenay2013classification}.
Robust algorithms for LNL do not focus on specific noisy instances but rather aim to design specific modules or mechanisms that allow networks to be well-trained despite the presence of label noise. These algorithms often employ techniques such as regularization, loss correction \cite{Patrini2016MakingDN}, and robust optimization to mitigate the effects of noise on the learning process \cite{zhang2018generalized}.

On the other hand, noise detection strategies aim at identifying and mitigating the impact of noisy data, thereby facilitating the training of more accurate models \cite{brodley1999identifying}.
Noise detection methods specifically target the erroneous labels within the dataset. These methods typically involve two stages: noise identification and data cleansing or reweighting. By accurately identifying noisy instances, these strategies enable the exclusion or correction of such data, thereby improving the overall quality of the training dataset \cite{song2022learning}.

We, on the contrary, focus on the correction of noisy labels after their detection. This is mainly due to the scientific application behind the idea, where each missed object is a potential discovery and valuable.

From another perspective, most of the existing methods focus on typical classification tasks and benchmarks, where the labels (and their respective noise) are of a categorical nature. \citet{Das2023UnderstandingSI,Santos2023ExpertisebasedWF} discuss linear regression in the context of Self Distillation, with a look at label noise. However, they do not cover more sophisticated models and/or non-categorical outputs.
\citet{Ponti2022ImprovingDQ} focus on tabular data and use training dynamics of Gradient Boosting Decision Trees.
Our application involves image generation (pixel-level regression) and is thus substantially different. We also test and show results on typical classification benchmarks and discuss how the original task is different, calling for a relatively more sophisticated technique.

A group of methods that rely on the specific state of the model throughout different stages of training, such as early stopping---\cite{Li2019GradientDW}. \citet{arpit2017closer} suggest that DNNs first learn simple patterns and subsequently memorize noisy data. \citet{liu2020early} suggest that deep neural networks, when trained on noisy labels, initially fit the data with clean labels during an ``early learning" phase and later begin to memorize the data with incorrect labels.

In contrast, we are network-state agnostic. Our method is, by design, able to learn the overfitting profiles, regardless of the stage at which we have stopped the training. This methodological choice is inspired by the fact that, in many real-world scenarios, one is not training a network from scratch, but fine-tuning a network already quite ``familiar'' with the task at hand. This also allows for the utilization of the technique in a closed-loop multi-cycle configuration, allowing the ecosystem to converge to the right answer.

Many studies have exploited the training dynamics of the models to tackle label noise. \citet{kohler2019uncertainty} detect noisy label data by analyzing the variations in predictive uncertainty distributions of a DNN between clean and noisy datasets. They use heuristically set rules to interpret the behavior of the curves, in their no-ground-truth setting. 
\citet{Jia2022LearningFT} explore training dynamics by training an LSTM, emphasizing the detection of label noise. While they highlight the idea of correction, they do not present a concrete correction algorithm.

\citet{tanaka2018joint} addresses the problem in the semi-supervised learning context, where one knows which data is labeled or not and only needs to assign pseudo-labels to unlabeled data. 
\citet{zhuo2022uncertainty} address the problem of noisy labels in the context of domain adaptation, by sample selection and reweighting.

MentorNet \cite{Jiang2017MentorNetLD}, Co-teaching \cite{Han2018CoteachingRT}, Co-teaching+ \cite{Yu2019HowDD} all use dual network architectures in which the two networks interact with each other during the training. They essentially focus on the (dis)agreements of the losses of the two networks for implicit sample selection. None of these focuses on the correction of the corrupted labels. 
\citet{shi2023choice} study the application of label noise detection in pediatric heart transplantation and rare disease detection.

\begin{figure*}[ht]
  \centering
  \includegraphics[width=\textwidth]{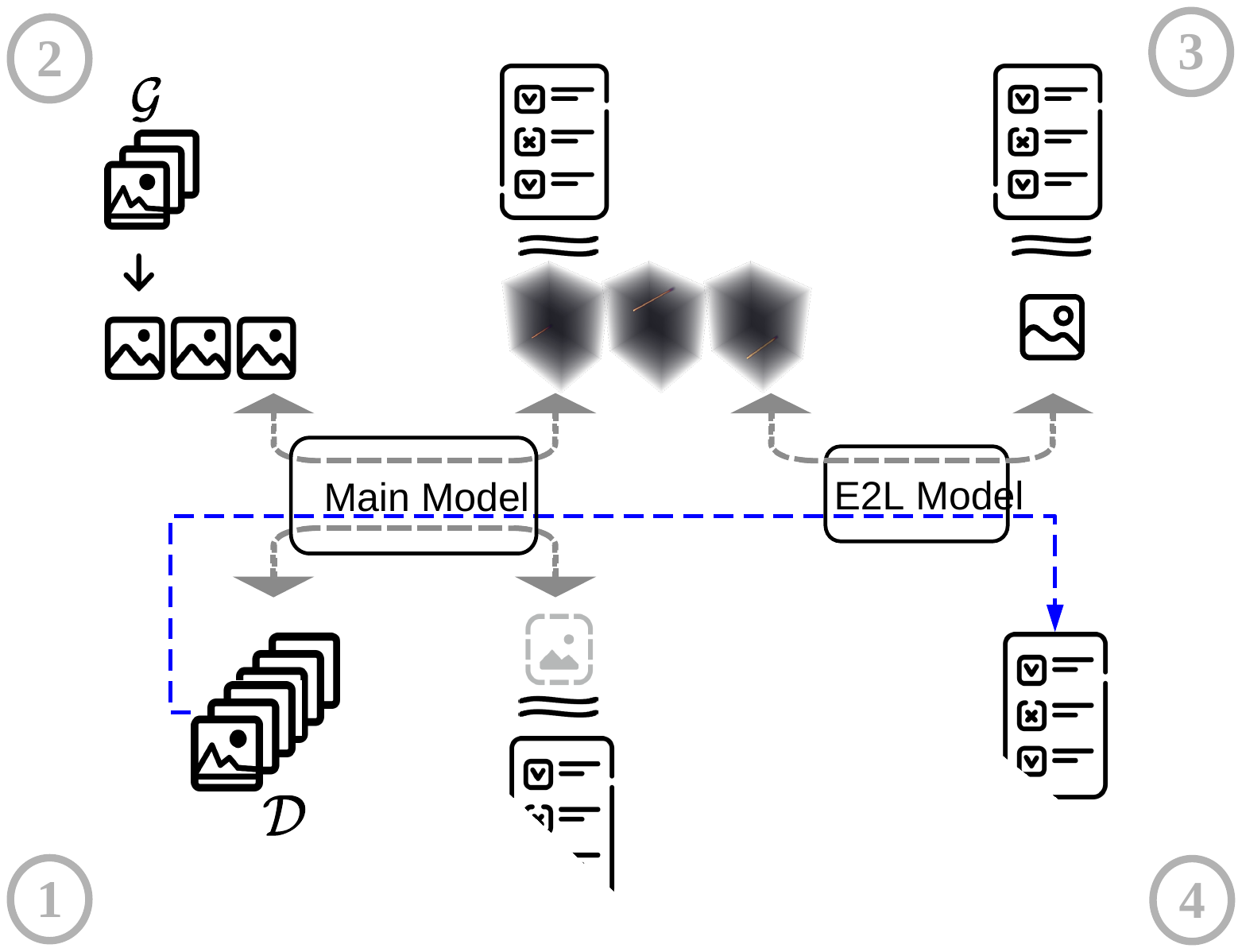}
  \caption{Illustration of various stages of a complete super-epoch. At step 1 (bottom left), the main model is trained on the training subset \(\mathcal{D}\) with the original noised labels. In step 2 (top left), individual samples from the gold subset \(\mathcal{G}\) are used to train the model to generate evolution cubes. During step 3 (top right), the E2L model is trained from scratch to learn to map evolution cubes of this super-epoch to clean labels. Finally, at step 4 (dashed blue arrow), the main subset \(\mathcal{D}\) is passed through the main and E2L models to give a cleaned-up version of the labels -- evolution cubes are generated on the fly.}
  \label{fig:your_label}
\end{figure*}

\textit{Dataset Cartography} \cite{Swayamdipta2020DatasetCM} uses training dynamics to characterize and diagnose datasets for natural language processing classification tasks. They leverage two main measures derived from training dynamics - confidence (mean probability of true label) and variability (standard deviation of true label probability) - to plot instances on a 2D map, revealing regions of easy-to-learn, hard-to-learn, and ambiguous examples. Their idea is quite close to the underlying concept of our method. However, our method does not need to capture the training dynamics of the network from scratch and starts capturing ``evolution history'' off a pre-trained state. Moreover, we do not stop at the detection of the label noise but emphasize correcting each of the erroneous labels as valuable elements of our special application.

\paragraph{Our contributions}
\begin{itemize}
    \item We use overfitting dynamics instead of training dynamics.
    \item We prioritize noisy label correction as a main objective.
    \item We are network state-agnostic: we do not assume any network states (early stopping, fully trained, etc.).
    \item We address label noise in image-like labels---the literature is almost always classification.
\end{itemize}

\begin{figure}[h]
  \centering
  \includegraphics[width=0.5\textwidth]{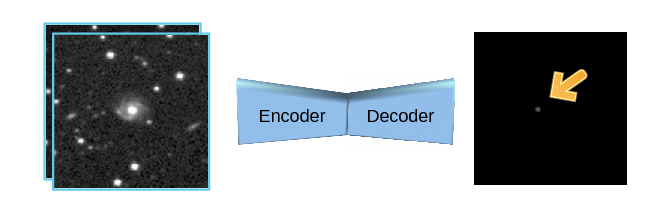}
  \caption{Image-based redefinition of the task of supernova detection. On the left, two images of the same region of the sky are passed to the network, and the output is defined as an image of the same size, containing only the reconstructed desired object \cite{sedaghat2018effective}}
  \label{fig:transinet}
\end{figure}
\section{Problem formulation}
Assume our dataset consists of two parts: a small 'gold subset' with clean labels, \(\mathcal{G}\), and a larger main subset, \(\mathcal{D}\), with possibly noised labels. For problem formulation, we proceed with \(\mathcal{D}\) alone and come back to \(\mathcal{G}\) for elaboration of the method in the next section.

Let \(\mathcal{D} = \{(\mathbf{x}_i, \tilde{y}_i)\}_{i=1}^N\) denote the dataset, where \(\mathbf{x}_i \in \mathbb{R}^d\) is the \(i\)-th input feature vector and \(\tilde{y}_i \in \mathcal{Y}\) is the corresponding noisy label. 
In this paper, the label space \(\mathcal{Y}\) is kept as flexible as possible. 
\(\tilde{y}_i\) is a sample from the noisy labels, which may not reflect the true underlying labels \(y_i^*\).

Label noise is often represented by \(P(\tilde{y}_i \mid y_i^*)\): the probability of observing \(\tilde{y}_i\) given the true label \(y_i^*\). 
A common model, in typical classification tasks, is the symmetric noise model where 
\begin{equation}
P(\tilde{y}_i = y_i^* \mid y_i^*) = 1 - \eta
\end{equation}
and 
\begin{equation}
P(\tilde{y}_i \neq y_i^* \mid y_i^*) = \frac{\eta}{C-1} \quad \forall \tilde{y}_i \, (\tilde{y}_i \neq y_i^*)\end{equation}
with \(\eta \in [0,1)\) representing the noise level and $C$ the number of classes. To keep the formulation as general as possible, we follow the same ``instance-independent'' formulation of label noise without any further assumptions, even though in the experiments we showcase the applicability of our method on datasets with more specific types of label noise too.

Let \(f(\mathbf{x}; \theta)\) be the deep neural network model parameterized by \(\theta\), which maps an input \(\mathbf{x}\) to an output \(\hat{y} = f(\mathbf{x}; \theta)\). The loss function, which can be applied to the entire training dataset or individual mini-batches, is defined as follows:
\[
\mathcal{L}(\theta) = \frac{1}{N} \sum_{i=1}^N \ell(f(\mathbf{x}_i; \theta), \tilde{y}_i),
\]
where \(\ell(\hat{y}, \tilde{y})\) denotes a chosen loss function that measures the discrepancy between the predicted label \(\hat{y}\) and the noisy label \(\tilde{y}\), and \(N\) represents the total count of samples in the dataset or mini-batch.
\section{Selfish Evolution: the method}
\label{sec:method}

\paragraph{Step 1 -- Initial training:}
The model \(f(\mathbf{x}; \theta)\) is trained on the main, noisy dataset \(\mathcal{D}\) for an arbitrary number of epochs:
\[
\theta = \arg\min_{\theta} \mathcal{L}(\theta; x, \tilde{y}).
\]

Let us refer to the intermittent state of the network after the interruption as 
\( \theta^{\dot{}} \)

\paragraph{Step 2 -- Overfitting and evolution:}
In this step, we resume training the model off \( \theta^{\dot{}} \), but continue training only on an individual sample to overfit. Mathematically, this can be expressed as:
\begin{equation}
\hat{\theta} = \arg\min_{\theta} \mathcal{L}(\theta; x_i, \tilde{y_i})
\end{equation}

which at each epoch (which consists of a single iteration), can be stated as:
\begin{equation}
\theta_i^t = \text{update}(\theta_{i}^{t-1}, \mathbf{x}_i, \tilde{y}_i), \quad t = 1, \ldots, T,
\end{equation}
\begin{equation}
y_i^t = f(\mathbf{x}_i; \theta_i^t), \quad t = 1, \ldots, T
\end{equation}

where \(T\) is the number of overfitting steps (epochs), and \(\theta_i^t\) represents the model parameters at step \(t\) for sample \(i\). This allows us to capture the ``evolution'' dynamics of the model:
\begin{equation}  
\mathcal{E}_i = \{y_i^t\}_{t=1}^T
\end{equation}

In our specific application where the labels are image-like tensors, \(\mathcal{E}_i\) are spatiotemporal, 3-D tensors and so are called ``evolution cubes''. 

We repeat the same overfitting process of this step for each of the samples in the subset at hand, indexed by \(i\). For each sample, we restart off the \( \theta^{\dot{}} \) state.

As we will show in the experiments, one can use a more generalized version of this step, where the overfitting target is not just a single sample, but a whole mini-batch, or a combination of them.
\paragraph{Step 3 -- Training of the Evolution-to-Label model:}
We train a secondary network \(g(\mathcal{E}; \phi)\) parameterized by \(\phi\) on these evolution cubes to detect and correct corrupted labels:
\[
\phi^* = \arg\min_{\phi} \ell(g(\mathcal{E}_i; \phi), y_i^*),
\]
where \(y_i^*\) are the true labels (or high-confidence corrected labels).

\subsection{Closed Loop Correction}
The trained secondary network, ``Evolution-to-Label'', maps the cubes, generated based on the current version of the labels, to a new set of labels--hopefully cleaner. We can optionally iterate the process in a closed-loop fashion, aiming for a mostly clean dataset:
\[
\tilde{y}_i^{(k+1)} = g(\mathcal{E}_i; \phi^{(k)}),
\]
where \(k\) is the iteration index. We use this iteration scheme in some of the simpler experiments of the next section, where each cycle is referred to as a ``super-epoch''.

\section{Experiments}
\subsection{Supernova detection}
Image-based supernova hunting is a pivotal task in astronomy. The de-facto way is to collect images of the same region of the sky, register, and co-add (average) them to get a template image. Then upon capturing each new image, a subtraction is performed, followed by noise removal, detection, etc.

\citet{sedaghat2018effective} redefine the task as an image generation task in which the output contains only the image of the supernova and nothing else -- \cref{fig:transinet}. ML-wise, it is close to the task of segmentation, in the sense that we essentially assign the value zero to pixels corresponding to unwanted objects. However, it is not just segmentation in the sense that the pixel values are not typical categorical values, but rather continuous scalars. It is also not a simple pixel-level regression task, in the sense that spatial coherence is important, especially in the presence of a supernova, where the shape needs to be preserved. The pixel values, at least in the original implementation of the method, represent the exact `flux'\footnote{A proxy of the apparent brightness of the object.} values of an ideally subtracted supernova. All these make it a spatiotemporal regression task. Something not often considered in typical label-noise research. In the below experiments, for the sake of comparability, we normalize all the target amplitudes, such that the idea output becomes a mere localization heat map.

\subsubsection{Data}We use data from the Dark Energy Science Collaboration (DESC) DC2 dataset: a simulated dataset covering a wide range of astrophysical phenomena with realistic simulations of the sky, containing billions of galaxies over a large area of the sky \cite{abolfathi2021lsst, abolfathi2021desc}. Our dataset consists of 3712 cutouts of size \(256\times256\) randomly centered around 373 unique supernovae. The relatively low number of images is, in part, chosen on purpose to emulate the challenging conditions of lack of labeled data in real-world astrophysical applications. 

\begin{figure*}[t]
  \centering
  \includegraphics[width=\textwidth]{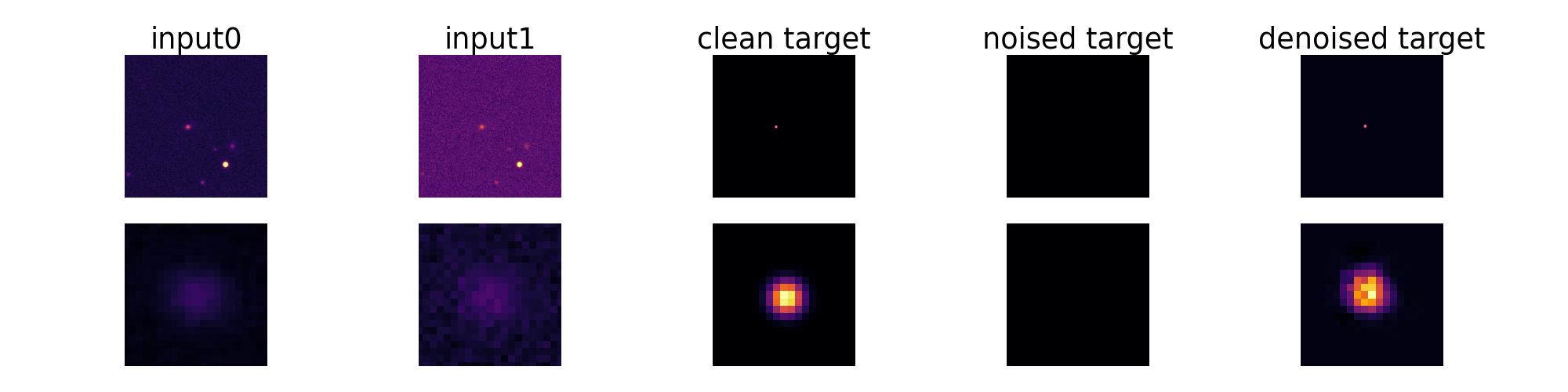}
  \caption{Results of denoising on one exemplar pair of inputs. The top row is the full image crop, while in the second row, we zoom in to have a clearer view of the target object. ``noised target'' is the blank target we have trained the primary network on. ``denoised target'' is the output of our algorithm, where the correct truth label is recovered.}
  \label{fig:result1}
\end{figure*}

\begin{figure*}[t]
  \centering
  \includegraphics[width=\textwidth]{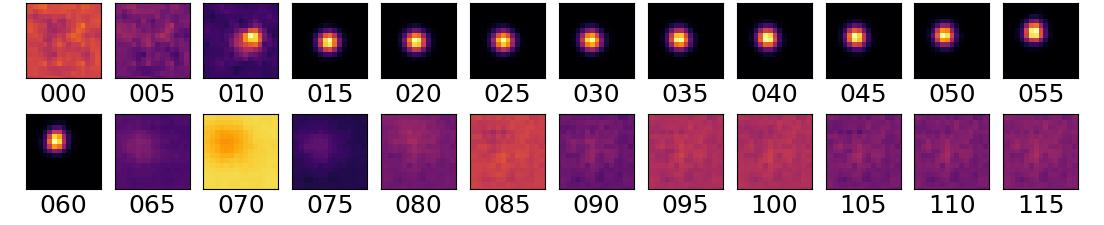}
  \caption{Exemplar illustration of a down-sampled, unrolled, evolution cube. 
  The first half (top row) is the first half of the evolution, where the network tries to overfit the support batch. In the second half, overfitting happens towards the single noised target. The race between the two overfitting schemes reveals subtle information about the clean label, which is exploited by our E2L model later on.
  }
  \label{fig:evolution}
\end{figure*}

\begin{table*}[h!]
  \caption{Noise correction quantitative results -- supernova detection}
  \label{sample-table}
  \centering
  \small
  \begin{tabular}{lcccc}
    \toprule
    Variant     & \small{Init. Clean Labels} & \small{Clean-Denoised Sim.} & \small{Clean-Denoised Sim. Hard.} & \small{Discovered objects} \\ 
    {} & ($\%$)  & (cosine,$ \%$) & ($\%$) & \\
    \midrule
    Baseline (full)          & 20.0  &  68.1 &  73.0 & - \\
    Selfish Evolution (full) & 20.0  &  75.6 & 82.7 & - \\
    Baseline (500)           & 50.0  &  8.9  &  0.0  & 0\\
    Selfish Evolution (500)  & 50.0  &  13.4 &  8.4  & 10 \\
    Baseline (full)          & 50.0  &  9.1  &  4.7 & 7 \\
    Selfish Evolution (full) & 50.0  &  31.8 &  \hl{50.1}  & \hl{817} \\
    \bottomrule
  \end{tabular}
  \label{tab:quant}
\end{table*}

We carefully split the dataset to prevent any object from leaking across subsets, resulting in $3205$ \textit{train} images and $507$ \textit{gold} samples. We also created several label-noised versions of the ground truth images: $20\%$, $50\%$, and $100\%$ noise.

\subsubsection{Initial training}
We use the exact same non-probabilistic, encoder-decoder architecture, as introduced in the original work of \citet{sedaghat2018effective}, to train the model on our training subset. We use a solver based on the ADAM optimizer \cite{kingma2014adam} and with an initial learning rate of $1e-4$.

\subsubsection{Evolution}
We use a mixed overfitting strategy to induce a race condition in the model dynamics: the model is pushed to overfit to a single clean mini-batch for the first half of the process. Then we switch the overfitting target to the single target label. The implementation consists of the below steps:
\begin{itemize}
    \item Initialize the main model with pre-trained weights.
    \item Pick one sample from the dataset (depending on the stage we are in)---the `Selfish Sample' hereafter.
    \item Pick a random batch from the clean dataset---the `Support Batch' hereafter.
    \item Initialize an empty cube.
    \item Continue training the model with the support batch, for a predefined number of epochs. 
    \item Infer on the Selfish Sample at the end of each epoch and append the output to the evolution cube.
    \item Switch to training of the model with the Selfish Sample for a predefined number of epochs.
    \item Infer on the Selfish Sample at the end of each epoch and append the output to the evolution cube.
    \item Start over -- includes reinitialization of the model with the pre-trained weights.
\end{itemize}

The last item is particularly important, since we want to capture comparable dynamics for each of the samples in the dataset, off of a fixed model state.
Also note that throughout the evolution process, regardless of which half we are in, there is only a single mini-batch involved. Therefore each epoch corresponds to a single iteration. When we are using the support batch, though, we need one extra forward pass with the Selfish Sample.

We use an ADAM solver \cite{kingma2014adam} with the parameters mentioned in \cref{tab:adam_hyperparams_updated}. The subtle difference between the two sections is due to the different behaviors we expect from the network: during tuning with the support batch, we want the gradients not to deviate too much from their last state, with the hope that in case of a noisy label, the model can lean towards the clean answer. In the second half, though, we want to allow the model to try to overfit to the `Selfish Sample'---\cite{mohammadi2020towards, keskar2017improving}.

\begin{table}[h]
  \centering
  \begin{tabular}{|c|c|c|}
    \hline
    \textbf{Hyperparameter} & \textbf{Support} & \textbf{Selfish} \\ \hline
    Learning Rate ($\alpha$) & 1e-4 & 1e-4 \\ \hline
    Weight Decay & 0.1 & 0 \\ \hline
    $\beta_1$ & 0.99 & 0.9 \\ \hline
    $\beta_2$ & 0.999 & 0.999 \\ \hline
  \end{tabular}
  \caption{Solver parameters used for the two parts of the evolution.}
  \label{tab:adam_hyperparams_updated}
\end{table}

For this experiment, we set the number of evolution epochs, $N_e$, to 60.

\begin{figure*}[h]
    \centering
        \includegraphics[width=0.7\textwidth]{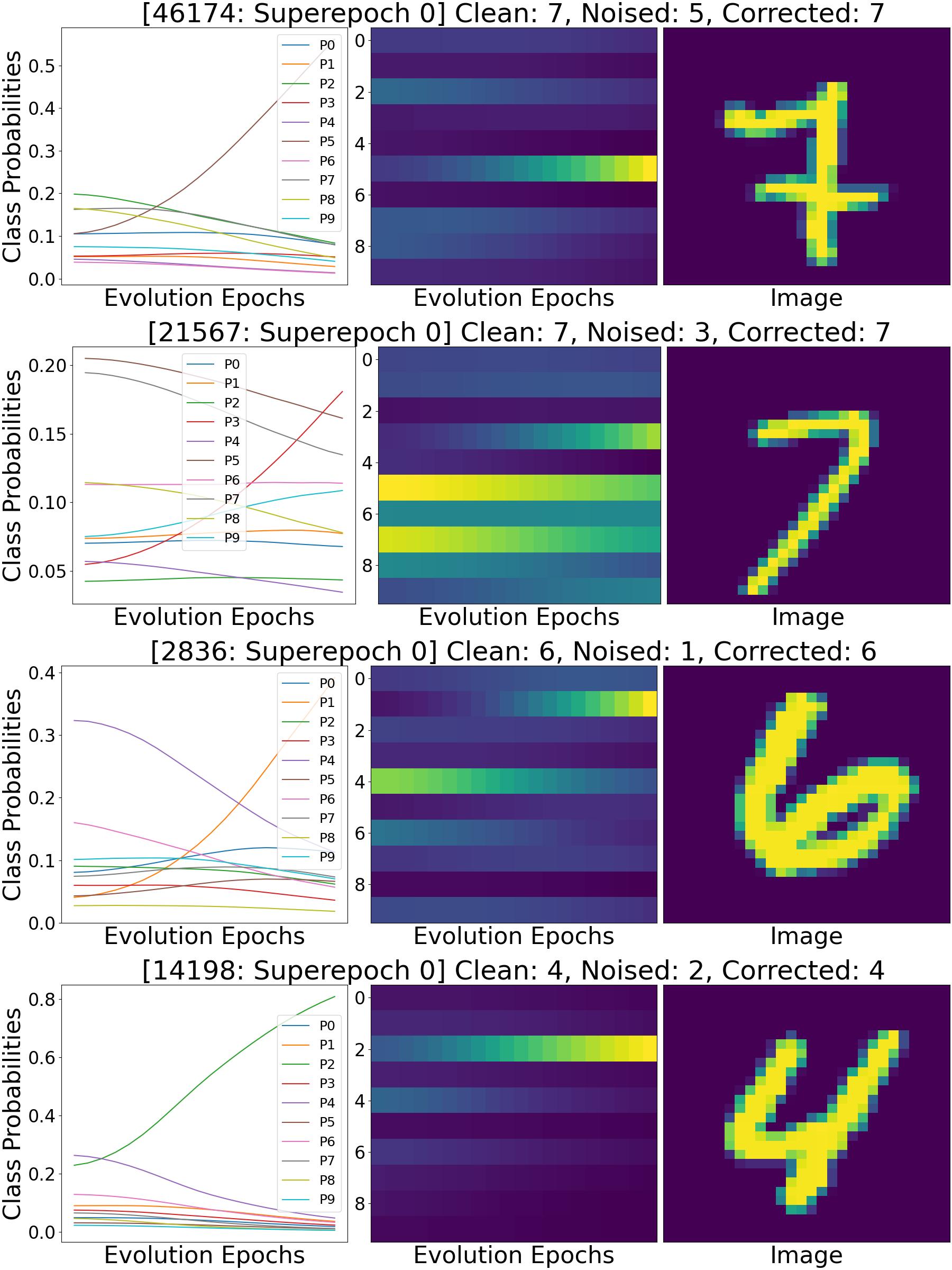}
    \caption{(left) Evolution histories of some exemplar noised samples in MNIST training set. (middle) Image-like presentation of the evolution histories. (right) The input images. Note how the clean label is not immediately distinguishable in the evolution patterns of the likelihoods---what our E2L model manages to exploit for inferring the clean label.}
    \label{fig:mnist_plot_stacked}
\end{figure*}
\subsubsection{Denoising}
For denoising, we do not separate the tasks of noise detection and correction. These steps take place implicitly and in conjunction with each other when we train a secondary model that directly maps the evolution cubes to clean labels; we refer to this model as the Evolution-to-Label mapper, or \texttt{E2L} in short.

Practically speaking, although the input data to \texttt{E2L} is a sequence in nature, it also matches the input to the main model -- it is only deeper. Therefore, given that the expected output is exactly of the same type, and to keep homogeneity in our implementations, we use the same architecture for \texttt{E2L} -- noting that a sequence-based model, like an RNN, may well replace our implementation. 

Moreover, since the number of gold samples, whose cubes are used for training of \texttt{E2L}, is too small, we use a `thinner' version of TransiNet with only half of the output channels in each hidden layer to avoid overfitting. We also use other regular measures such as image flipping (with a $50\%$ chance in each image dimension) and shifting (with a uniform probability between 0 and 20 pixels in each image dimension) during training.

\texttt{E2L} is trained on the evolution cubes obtained from the gold dataset. It is indeed trained on a combination of two versions of it: cubes from the clean version, and cubes from a $100\%$ noised version. This way we try to maximize the types of the evolutions \texttt{E2L} sees, even those that do not need to be corrected!

\subsubsection{Results}
We pass the evolution cubes through the \texttt{E2L} network and infer estimates for the corrected labels. We define and compute multiple evaluation metrics: 
\begin{itemize}
    \item a soft similarity metric: simple cosine distance between clean and denoised, 
    \item a hard similarity metric: thresholded version of the soft similarity metric.
    \item Discovery rate/count: the number of recovered objects (above threshold).
\end{itemize}

As stated throughout the paper, our main objective is \textit{label noise correction}. Therefore, unlike many studies, we do not evaluate the performance on a clean validation set, but directly on the training set. \cref{tab:quant} summarizes the quantitative results. We ran several experiments with various noise levels and hyperparameters, but only bring the three main representative ones in the table. `Baseline' is the output of the primary network, directly trained on the noised dataset---no correction. The setup designated by `500' is one in which we set extremely hard conditions by only using the first 500 samples from the training set. In contrast, in the `full' version, we used all the training samples. We recovered 817 supernovae that were previously missed!

\Cref{fig:result1,fig:evolution} depict how the evolution cube and the corrected label look like in an exemplar case.

\subsection{Standard image classification---MNIST}

We test our method on MNIST~\cite{lecun-mnisthandwrittendigit-2010}, mainly to illustrate the underlying mechanisms of our proposed method, in a more manageable application. We prepare three datasets by modifying the labels of the MNIST dataset: (1) the clean dataset identical to the original MNIST dataset, (2) the noised dataset whose 80\% of its labels have been randomly changed, and (3) the noised dataset whose all of the labels have been randomly altered. We then separate the 60,000 images of the train MNIST dataset into two groups, the first 51,000 being the training set and the last 9,000 images being the gold set.

\begin{table*}[h!]
  \caption{Noise correction quantitative results -- MNIST}
  \label{mnist_results_1}
  \centering
  \small
  \begin{tabular}{lcccc}
    \toprule
    Variant     & \small{Init. Noise Levels} & \small{Clean Percentage}  & \small{Final Noise Levels} & \small{Learnable Parameters} \\ 
    {} & ($\%$)  & ($\%$) & ($\%$) & \\
    \midrule
    Baseline                  & 50.0  & 91.3  & 8.7 & 431,080    \\
    Selfish Evolution (1 super-epoch)                 & 50.0  & 89.7 & 10.3 & 431,080  \\
    Selfish Evolution (10 super-epochs)             & 50.0  & 93.9 & 6.1 & 431,080  \\
    \midrule
    Baseline                                              & 80.0  &  58.7  & 41.3  & 431,080   \\
    Selfish Evolution (1 super-epoch)                          & 80.0  &  63.6 & 36.4 & 431,080 \\
    Selfish Evolution (10 super-epochs)                         & 80.0  &  78.8 & 21.2 & 431,080 \\
    Co-teaching                                         & 80.0 & 78.3 & 21.7 & 4,432,266 \\
    \bottomrule
  \end{tabular}
\end{table*}

We use the train set of the partially noised dataset to train the primary network, after which we feed each image in the gold set into the model individually. In other words, we continue training the trained primary model using only a single image from the gold set. For each forward pass in a single iteration, we record the output. At the end of the training, we obtain an evolution history, i.e., a temporal strip (i.e., cube) of the likelihood of the prediction over iterations. We then reset the primary model to its original state before we fed another image from the gold set. The procedure is repeated for every image in the set. Thus, we obtain another dataset consisting of evolution histories corresponding to each image in the gold set. In this step, this procedure is performed for (1) the gold clean subset, and (2) the gold noisy subset. Thus, we obtained two sets of evolution histories.

In Figure \ref{fig:mnist_plot_stacked}, we can see the class probability corresponding to the noised label increases over the iterations. For example, the image index 46174 has a clean label of ``7'' and a noised label of ``5.'' Over the iterations, the model predicts that the image is ``5'' as we assigned its noised label to be ``5'' although it initially predicted that ``2'' is most likely in the first iteration.

\begin{figure*}[h]
    \centering
        \includegraphics[width=\textwidth]{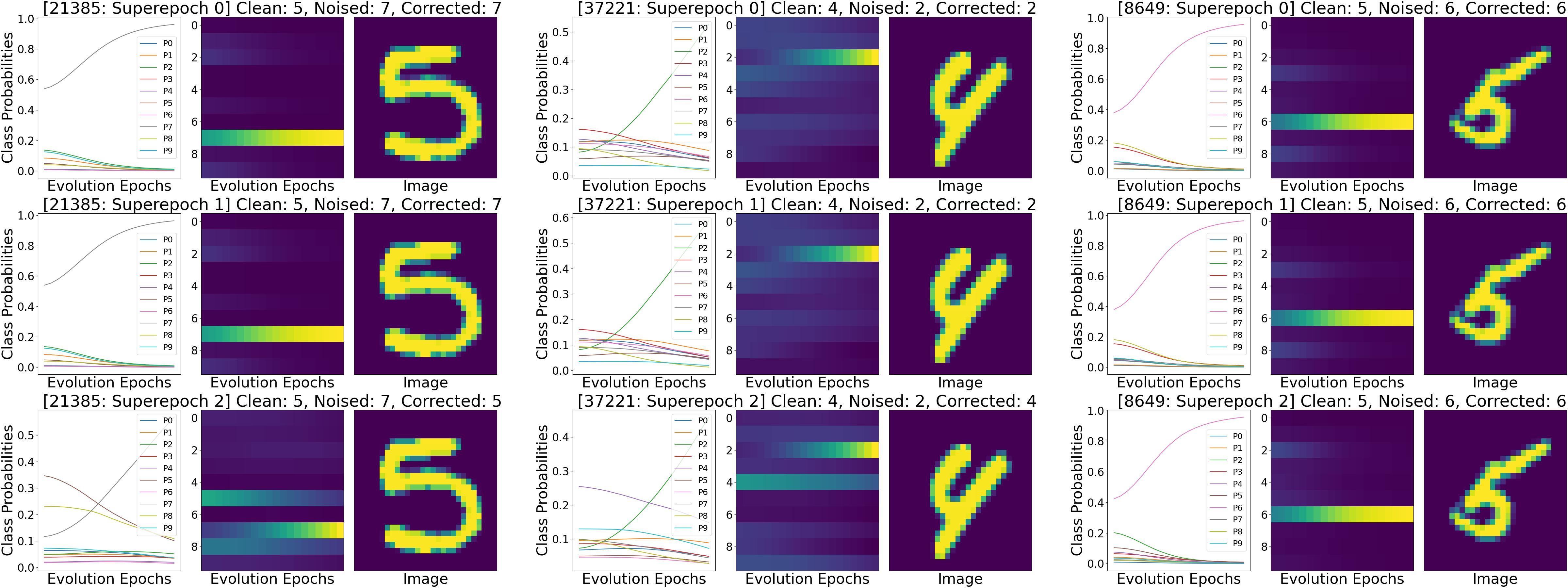}
    \caption{Exemplar evolution histories of noised data points in the MNIST train set. The first and second columns show that the labels are corrected only after a few iterations. The third column shows a failed example.}
    \label{fig:mnist_plot_success_failure}
\end{figure*}

The gold evolution histories are combined and fed into the secondary network (E2L). Instead of images and their corresponding labels, the dataset is the evolution history, and the target is the \textit{clean} label of the original image of the evolution history. The gold clean evolution histories can be fed into the model directly. However, in gold noisy evolution histories, the noised labels were corrected before being fed into the model.

After training the E2L model, we evolved the \textit{train} subset of the MNIST train set (i.e., the 51,000 images with partially noised labels) and obtained their evolution histories. Then, we fed these train subset evolution histories into the trained E2L model. The output of this stage would be used to update the original noised dataset from the first step.

The two models we utilized in this MNIST experiment are LeNet\cite{lecun-mnisthandwrittendigit-2010}. This LeNet model consists of 2 CONV-RELU-POOL layers, 2 fully connected layers, and one softmax. 

The results on MNIST are shown in Table \ref{mnist_results_1}. We ran experiments on 50\% and 80\% noised datasets. The results consist of 10-super-epoch runs and 1-super-epoch runs in comparison with baselines. Our baseline is the performance of the primary model itself. Additionally, we also include the performance of the Co-teaching algorithm for comparison purposes. Note though that this method is originally evaluated on a clean validation set. For us to be able to make a fair comparison, we ran the noisy training set through the final version of the trained model. Even though Co-teaching uses a much larger network architecture, we still perform on par.

\section{Discussion and future work}
We introduced the novel idea of detecting and correcting noisy labels based on overfitting dynamics. Apart from its novelty, the proposed method helped us recover (discover) more than fifty percent of the missed supernovae in an exemplar dataset, which is beyond significant in the field of astronomy. We make the source code and the supernova dataset available to the public upon acceptance of the paper.
Furthermore, the method has the potential for utilization in domain-adaptation scenarios: a dataset from another domain with all-blank labels is a perfect fit for the algorithm.
Although we were focused on the specific task, we showcased the efficiency of the method on the rather typical classification task. We showed that the mere use of the `Selfish' part of the evolution suffices in the case of this simple task. We bring the results of the same experiments on the CIFAR \cite{krizhevsky2009learning} dataset in the supplementary material.
\newpage

\bibliographystyle{plainnat}
\bibliography{ref}

\newpage

\end{document}